\documentclass{article} % For LaTeX2e
\usepackage{iclr2026_arxiv,times}

\usepackage{amsmath,amsfonts,bm}

\def\eqref#1{equation~\ref{#1}}
\def\1{\bm{1}}

\DeclareMathAlphabet{\mathsfit}{\encodingdefault}{\sfdefault}{m}{sl}
\SetMathAlphabet{\mathsfit}{bold}{\encodingdefault}{\sfdefault}{bx}{n}

\usepackage{hyperref}
\usepackage{url}
\usepackage{graphicx}
\usepackage{placeins}


\title{Muon as a Residual Connection}

\author{Hao Huang \\ College of Computer Science and Technology \\ Zhejiang University \\ Hangzhou, Zhejiang, China \\ \texttt{goodhao@zju.edu.cn} }

\iclrfinalcopy % arXiv/preprint version: show author information
\begin{document}

\maketitle

\begin{abstract}
Muon has recently emerged as one of the most effective optimizers for training large neural networks, yet its empirical success has been explained from several different perspectives. In this paper, we propose a simple mechanistic interpretation: Muon can be understood as an implicit residual connection during training. Specifically, orthogonalizing the update can sacrifice some immediate gradient fidelity while improving representation preservation for downstream layers. We study this trade-off in controlled linear optimization settings, where Muon can learn representations that are slower to fit a local target but easier for downstream layers to exploit. Our results suggest a conceptual explanation for Muon and a design perspective for optimizers that balance local descent with downstream usability.\footnote{Code is available at \url{https://github.com/huanghao-sss/muon_interpretation}.}
\end{abstract}

\section{Introduction}

In recent years, Muon has attracted considerable attention as a highly effective optimizer for deep neural networks, delivering state-of-the-art performance on a wide range of architectures, from CNNs to LLMs.

Muon is built upon a simple idea---orthogonalizing the updates of matrix parameters. Yet, despite its simplicity, its empirical success has inspired a surprisingly diverse collection of sophisticated explanations, including spectral steepest descent, sharpness, curvature, and long-tail learning.

A natural question arises before introducing yet another explanation of Muon: why does Muon need a new explanation at all?
Rather than viewing these explanations as mutually exclusive, we begin from a broader perspective on scientific explanation. As emphasized in the philosophy of science, a scientific phenomenon often admits multiple valid explanations, each highlighting different causal mechanisms, levels of abstraction, or explanatory purposes.

Different explanations are valuable not only because they improve our understanding of a phenomenon, but also because they inspire different directions for future research. For example, interpreting Muon through the lens of spectral-norm steepest descent naturally motivates the exploration of optimizers based on alternative matrix norms. In contrast, explaining Muon's success through curvature suggests incorporating richer second-order information into optimizer design. From this perspective, a new explanation is valuable not merely as another interpretation, but as a new source of hypotheses and algorithmic design principles.

Our goal is therefore not to replace existing explanations, but to introduce a new explanatory perspective that complements them.
Specifically, we argue that Muon can be understood as an implicit residual connection, providing a simple and intuitive framework for reasoning about how optimization can affect downstream representation usability.

\section{Background}

\subsection{The Muon Optimizer}

Muon is a recently proposed optimizer that has attracted considerable attention due to its remarkable empirical success, from benchmarks such as NanoGPT and CIFAR-10 to frontier large language models including Kimi K2 and DeepSeek V4.

At its core, Muon replaces the standard parameter update
$W \leftarrow W + \Delta W$
with an orthogonalized update
$W \leftarrow W + \operatorname{Orth}(\Delta W)$,
where $\operatorname{Orth}(\Delta W)$ denotes the orthogonal matrix closest to $\Delta W$.

\subsection{Existing Interpretations of Muon}

Existing interpretations of Muon emphasize several complementary aspects of its behavior. The spectral-norm steepest-descent view interprets the orthogonalized update as the steepest descent direction under the matrix spectral norm~\citep{bernstein2024old}, while the trust-region view derives the Muon step as the exact solution of a non-Euclidean trust-region subproblem~\citep{kovalev2025trustregion}. Other work views Muon with decoupled weight decay as a stochastic Frank--Wolfe method on a spectral-norm ball, thereby emphasizing the constrained optimization trajectory rather than a single-step update direction~\citep{sfyraki2025lions}. A further line of work connects Muon to rare directions and long-tail learning, showing that spectral normalization can prevent dominant gradient modes from overwhelming weaker but important directions~\citep{vasudeva2025spectral,wang2025tailassociative}. Curvature-based explanations argue that Muon can achieve comparable first-order progress to Adam while incurring a smaller second-order curvature penalty~\citep{wang2026curvature}. The spectral Wasserstein-flow perspective places Muon within a continuous-time optimal-transport framework for spectrally normalized optimization~\citep{peyre2026spectralwasserstein}. Finally, another interpretation emphasizes batch gradient alignment and directional descent potential, arguing that exact orthogonalization is not uniquely responsible for Muon's behavior~\citep{shumaylov2026muon}. We provide a more detailed summary of these interpretations in Appendix~\ref{app:existing-interpretations}.

These existing interpretations offer useful perspectives. Nevertheless, we seek an explanation of a different nature---one that is more intuitive and accessible, especially for readers without extensive mathematical training. Rather than invoking external mathematical frameworks, we aim for a mechanistic account grounded solely in common knowledge and basic principles familiar to the deep learning community. Specifically, we investigate the hypothesis that Muon can be understood as an implicit residual connection, reducing the question of why Muon works well to the question of why residual connections work well.

\subsection{Residual Connections in Deep Learning}

Residual connections have become one of the most fundamental architectural components in modern deep learning since their introduction in ResNets~\citep{he2016resnet}.
The basic idea is to add the input of a block directly to its output:
\[
x_{l+1}=x_l+\text{Block}(x_l),
\]
where the identity shortcut carries $x_l$ forward unchanged while $\text{Block}$ applies the usual nonlinear transformation.
This simple modification substantially alleviates optimization difficulties in very deep networks and has enabled the successful training of models with hundreds or even thousands of layers~\citep{he2016resnet}.

Residual connections have been studied from a variety of theoretical perspectives.
Existing explanations include identity-mapping views, which argue that residual blocks make it easier for layers to represent or stay close to the identity mapping~\citep{he2016identity}; path-ensemble views, which interpret ResNets as ensembles of many paths of different effective depths and attribute their trainability to short gradient-carrying paths~\citep{veit2016residual}; and dynamical-systems views, in which very deep residual networks are regarded as discretizations of ordinary differential equations~\citep{chen2018neural}.

Today, residual connections constitute a core design principle across convolutional networks and Transformers. In this paper, we emphasize a mechanistic role of residual connections: they pass both $x_l$ and $\text{Block}(x_l)$ to downstream layers. In other words, downstream layers need not rely exclusively on the transformed representation, but can choose, mix, ignore, or amplify information from the preserved upstream representation and the newly transformed representation~\citep{longon2024residualstream}.

\section{Muon as an Implicit Residual Connection}

\subsection{Orthogonal Updates as Residual Connection}

Residual connections explicitly transform the representation as
\[
x \leftarrow x+\text{Block}(x).
\]
When $\text{Block}$ is a linear layer, this is equivalent to
\[
x \leftarrow (I+W)x,
\]
or, equivalently,
\[
W \leftarrow W+I.
\]
More generally, the identity matrix is not essential: any orthogonal matrix $R$ can equally serve as the skip connection,
\[
W \leftarrow W+R.
\]

During optimization, the weight matrix evolves according to
\[
W \leftarrow W+\Delta W.
\]
Muon replaces the update by its orthogonalized counterpart,
\[
W \leftarrow W+\operatorname{Orth}(\Delta W),
\]
where $\operatorname{Orth}(\Delta W)$ denotes the nearest orthogonal matrix to $\Delta W$.

From this perspective, residual connections explicitly add an orthogonal transformation to the network weights, whereas Muon implicitly injects an orthogonal transformation through the optimization trajectory.

This is intended as a mechanistic explanation rather than a strict mathematical equivalence between Muon and residual architectures. The key analogy is that both mechanisms create an option for downstream computation. A residual block explicitly exposes both the preserved upstream representation and the transformed representation to later layers. Muon, by contrast, creates this option through the training dynamics: the update must still serve as a descent direction for the current parameter, but its orthogonalized component also tends to preserve and reshape the representation in a way that remains easy for downstream layers to use. Thus, Muon's update can be understood as serving two coupled roles: preserving gradient fidelity for the current optimization step while promoting representation preservation for downstream layers.

\subsection{The Gradient Fidelity--Representation Preservation Trade-off}

Orthogonalizing the update inevitably modifies its original direction. On the one hand, remaining closer to the raw update preserves a larger component along the gradient direction, leading to a greater first-order decrease in the objective, namely higher \emph{gradient fidelity}. On the other hand, as discussed in the previous section, orthogonal updates implicitly inject residual transformations into the network weights, allowing representations learned by earlier layers to be propagated to deeper layers with less distortion and resulting in stronger \emph{representation preservation}. Muon can therefore be viewed as balancing these two competing objectives.

This perspective is complementary to previous trade-off interpretations of Muon. For example, recent work argues that Muon balances batch gradient alignment and directional descent potential~\citep{shumaylov2026muon}. In contrast, we interpret Muon as balancing \emph{gradient fidelity}, which favors immediate loss reduction, and \emph{representation preservation}, which favors preserving representations for subsequent layers through residual transformations.

\section{A Two-Phase Linear Model}

\subsection{Problem Setup and Motivation}

Gradient backpropagation is the standard paradigm for training deep neural networks. It decomposes the optimization of the overall loss into layer-wise subproblems by propagating gradients through the chain rule, indicating how changing the parameters of each layer affects the final objective. An alternative paradigm is target propagation~\citep{lee2015difference}, which directly assigns each layer a desired target representation rather than a gradient direction. Although the optimization mechanisms differ, both approaches share the same underlying philosophy: decomposing a global optimization problem into a collection of local subproblems.

Our experiment is motivated by this viewpoint, although we do not implement target propagation itself. Instead, we use it as a conceptual framework for analyzing different optimization strategies. Suppose a target transformation $A$ is assigned to a linear layer with weight matrix $W$. The local optimization problem is
\[
W \rightarrow A,
\]
which is achieved through the additive update
\[
W \leftarrow W+\Delta W.
\]

Under this local objective, SGD follows the steepest descent direction and is therefore expected to approach $A$ faster than Muon, whose update is first orthogonalized before being applied. If this local objective were the only objective of interest, SGD would naturally be preferred.

This immediately raises two questions. Is the local objective itself the one that ultimately matters? Even if it is, must it necessarily be realized by the current layer?

To answer these questions, consider two consecutive linear layers with weights $W_1$ and $W_2$. Realizing the target through the current layer corresponds to the additive update
\[
W_1 \leftarrow W_1+\Delta W_1.
\]
Alternatively, the same target transformation may be realized by adapting the downstream layer so that
\[
W_2W_1 \rightarrow A.
\]
Updating the second layer,
\[
W_2 \leftarrow W_2+\Delta W_2,
\]
induces
\[
\Delta(W_2W_1)
\approx
(\Delta W_2)W_1,
\]
which corresponds to a multiplicative update of $W_1$.

This suggests that the layer-wise objective can be redistributed: part of the additive update that would otherwise be performed by the current layer can be amortized into multiplicative updates in downstream layers. Motivated by this observation, we present an experiment to demonstrate one possible mechanism: although Muon performs slower additive updates than SGD at the current layer, it can induce faster downstream multiplicative updates, leading to faster overall optimization than SGD.

We consider a two-layer linear network
\[
\hat y=W_2W_1x,
\]
where the target transformation is a Gaussian random matrix
\[
A\in\mathbb R^{d\times d},
\qquad
A_{ij}\overset{\mathrm{i.i.d.}}{\sim}\mathcal N(0,1/d),
\qquad
d=64.
\]

We compute the singular value decomposition
\[
A=U\Sigma V^\top,
\]
and construct one valid factorization
\[
W_1^\star=\sqrt{\Sigma}V^\top,
\qquad
W_2^\star=U\sqrt{\Sigma},
\]
which satisfies
\[
W_2^\star W_1^\star=A.
\]
The first factor $W_1^\star$ serves as the local target of the first layer, while the overall objective remains realizing the global transformation $A$.

The experiment consists of two phases.

In \textbf{Phase~1}, we freeze $W_2$ and optimize $W_1$ toward the local target
\[
\min_{W_1}
\frac12\|W_1-W_1^\star\|_F^2.
\]
We compare SGD and Muon under identical learning rates and optimization budgets.

In \textbf{Phase~2}, the learned $W_1$ is frozen, while $W_2$ is randomly reinitialized and optimized using SGD in both settings by minimizing
\[
\min_{W_2}
\frac12\|W_2W_1-A\|_F^2.
\]
Since Phase~2 uses exactly the same optimizer, hyperparameters, and initialization strategy in both experiments, the only difference is the representation learned by $W_1$ during Phase~1. Consequently, any difference in convergence speed directly reflects how easily the learned representation can be exploited by downstream layers.

The executed experiment uses $d=64$. Phase~1 runs for $600$ steps with learning rate $0.2$. Phase~2 uses SGD with learning rate $0.02$ and stops when
\[
\frac{\|W_2W_1-A\|_F}{\|A\|_F}\le 0.05.
\]
Distances in Table~\ref{tab:two-phase-results} are normalized by $\|A\|_F$.

\begin{table}[t]
\centering
\begin{tabular}{c|r r r r r}
Phase~1 optimizer & $\|W_1-W_1^\star\|_F$ & $W_1$ flatness & Phase~2 steps & total steps & final error \\
\hline
SGD & 0.0312 & 0.4301 & 13436 & 14036 & 0.049999 \\
Muon & 0.0447 & 0.4389 & 13321 & 13921 & 0.049999
\end{tabular}
\caption{Two-phase linear experiment. Distances are normalized by $\|A\|_F$. Spectrum flatness is $\|W_1\|_F^2/\|W_1\|_2^2$ divided by $d$ (an effective-rank proxy; higher means a flatter singular spectrum). SGD reaches a smaller Phase~1 local error, while Muon's $W_1$ is flatter and requires fewer downstream SGD steps in Phase~2.}
\label{tab:two-phase-results}
\end{table}

\subsection{Phase 1: Anchoring $W_1$}

Phase~1 intentionally favors SGD by evaluating only the local objective
\[
L_1(W_1)
=
\frac12
\|W_1-W_1^\star\|_F^2.
\]
Since SGD follows the steepest descent direction of this objective, whereas Muon first orthogonalizes the update before applying it, we expect SGD to approach $W_1^\star$ more rapidly.

The result in Table~\ref{tab:two-phase-results} confirms this expectation. After the same $600$ Phase~1 steps, SGD reaches a normalized reconstruction error of $0.0312$, while Muon reaches $0.0447$. Thus Muon sacrifices local optimization efficiency on this layer-wise subproblem.

\subsection{Phase 2: Composing $W_2W_1=A$}

Phase~2 evaluates whether the representation learned during Phase~1 facilitates downstream optimization. The learned $W_1$ is frozen, while $W_2$ is optimized from scratch by minimizing
\[
L_2(W_2)
=
\frac12
\|W_2W_1-A\|_F^2.
\]

Unlike Phase~1, both optimization paths now use exactly the same SGD optimizer. Therefore, any difference in convergence speed must originate solely from the representation encoded by $W_1$, rather than from differences between optimization algorithms.

Interestingly, although Muon converges more slowly in Phase~1, it requires fewer optimization steps in Phase~2. With SGD-trained $W_1$, the downstream layer needs $13436$ steps to reach the threshold; with Muon-trained $W_1$, it needs $13321$ steps, saving $115$ downstream steps. This indicates that the representation learned by Muon is easier for the downstream layer to utilize despite being farther from the local target $W_1^\star$.

\subsection{Why Orthogonal $W_1$ Accelerates Phase 2}

The behavior observed in Phase~2 admits a simple theoretical explanation. With $W_1$ fixed, the downstream objective is
\[
L(W_2)
=
\frac12
\|W_2W_1-A\|_F^2,
\qquad
\nabla_{W_2}L
=
(W_2W_1-A)W_1^\top .
\]
Letting $E_t=W_{2,t}W_1-A$, a single SGD step implies
\[
E_{t+1}
=
E_t
(I-\eta W_1^\top W_1).
\]
Thus, downstream convergence is governed by the spectrum of $W_1^\top W_1$, rather than only by how close $W_1$ is to the local target $W_1^\star$; see Appendix~\ref{app:phase2-convergence} for the singular-mode derivation. A flatter singular spectrum and smaller condition number make the downstream problem better conditioned. This explains how Muon can sacrifice local optimization accuracy in Phase~1 while enabling faster downstream optimization in Phase~2.

The same spectral argument also connects this experiment to linear probing: a better-conditioned fixed representation makes downstream linear prediction easier. We discuss this connection in Appendix~\ref{app:linear-probing}.

\section{\texorpdfstring{$\tau$}{tau} Scheduling}

\subsection{From Joint Updates to Segment Alternation}

The two-phase experiment above deliberately separates representation learning from downstream adaptation. We next ask whether the same trade-off remains visible when both layers are trained end-to-end on the composition loss. To this end, we use a simple $\tau$ schedule.

Consider the same two-layer linear model
\[
\hat y=W_2W_1x,
\]
optimized directly using the composition loss
\[
\widetilde L(W_1,W_2)
=
\frac{\|W_2W_1-A\|_F^2}{\|A\|_F^2}.
\]

The parameter $\tau$ controls how long the other layer is frozen:
\[
\begin{cases}
\tau=0, & \text{joint updates of }W_1\text{ and }W_2\text{ at every step};\\
\tau\ge 1, & \tau\text{ steps updating only }W_1\text{, followed by }\tau\text{ steps updating only }W_2.
\end{cases}
\]
Here $\tau$ sets the segment length of the alternating updates. Larger $\tau$ makes the $W_1$-only and $W_2$-only phases more pronounced.

Throughout this section, both methods optimize the same composition loss from the same initialization. The downstream layer $W_2$ is always updated using SGD. The only difference is the optimizer used for $W_1$: the SGD path uses SGD for both layers, while the Muon path uses Muon for $W_1$ and SGD for $W_2$.

\subsection{Empirical Behavior of the \texorpdfstring{$\tau$}{tau} Schedule}

We run the sweep with $d=64$, learning rate $0.02$, and stop when
\[
\frac{\|W_2W_1-A\|_F}{\|A\|_F}\le 0.05.
\]
The number of optimization steps required to reach this threshold is reported below.

\begin{center}
\begin{tabular}{c|r r r}
schedule & SGD on $W_1$ & Muon on $W_1$ & steps saved \\
\hline
joint $(\tau=0)$ & 13054 & 11535 & 1519 \\
$\tau=1$ & 26109 & 23064 & 3045 \\
$\tau=10$ & 26122 & 23077 & 3045 \\
$\tau=50$ & 26165 & 23064 & 3101 \\
$\tau=200$ & 26221 & 23352 & 2869 \\
$\tau=600$ & 26553 & 21975 & 4578
\end{tabular}
\end{center}

\begin{figure}[!htbp]
\centering
\includegraphics[width=\linewidth]{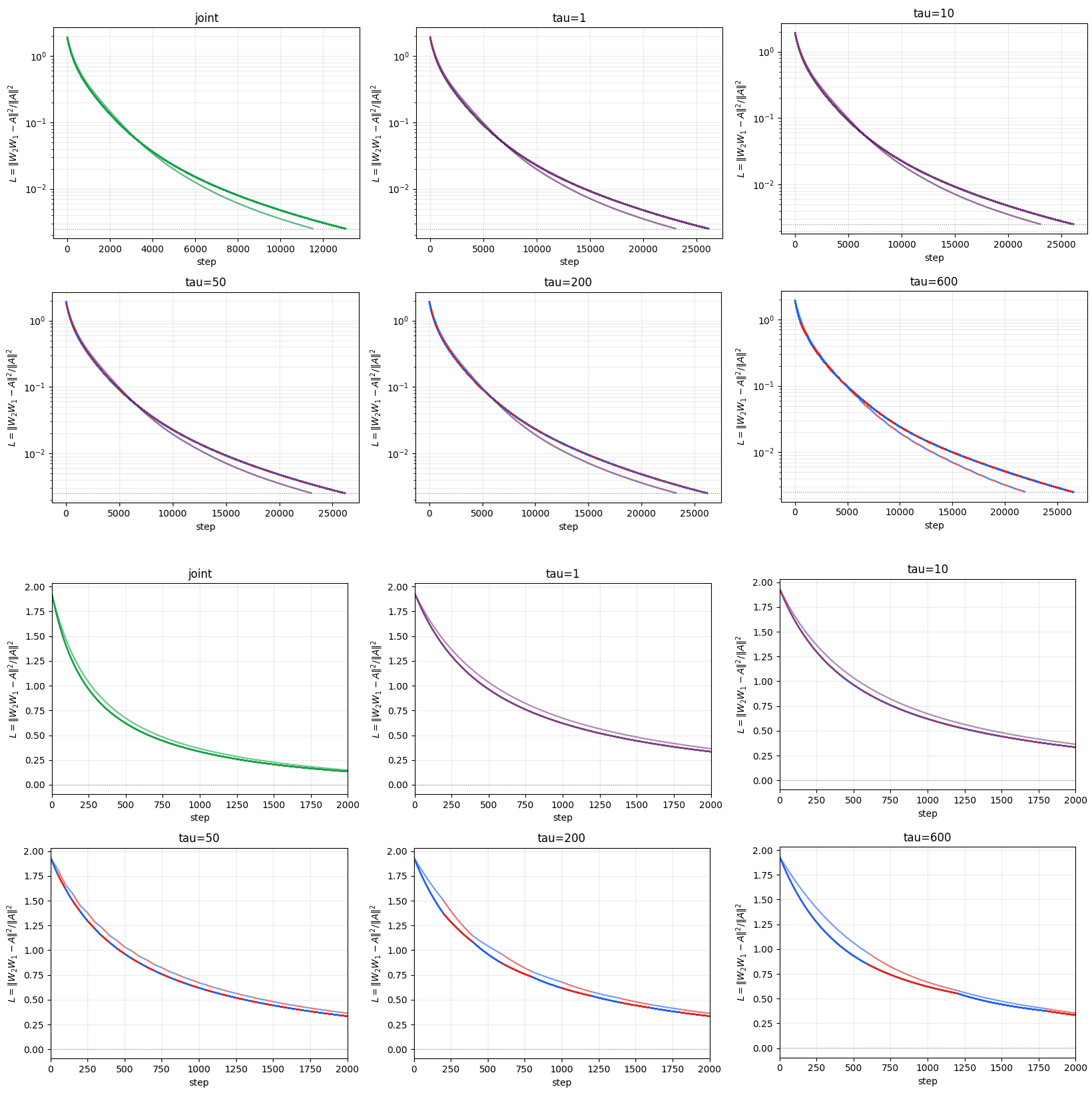}
\caption{Composition loss under the $\tau$ schedule. Top: full trajectories on a logarithmic scale. Bottom: early-step zoom on a linear scale. Solid curves use SGD for $W_1$, while dashed curves use Muon for $W_1$; $W_2$ is always trained with SGD. Colors indicate which layer is updated at each step: joint updates in green, $W_1$-only segments in blue, and $W_2$-only segments in red.}
\label{fig:tau-loss-sweep}
\end{figure}

Muon reaches the target loss faster for every schedule tested, including the fully joint case $\tau=0$; the full log-scale trajectories and early linear-scale behavior are shown together in Figure~\ref{fig:tau-loss-sweep}. The alternating schedules require roughly twice as many total steps as joint training because only one factor is updated at each step, but the comparison within each schedule remains matched: the two paths differ only in the $W_1$ optimizer. The advantage of Muon is particularly visible for larger segment lengths, where the downstream effect of the learned representation has more time to appear within each $W_2$ segment.

\subsection{Segment-wise Dynamics of the \texorpdfstring{$\tau$}{tau} Schedule}

The colored segments in Figure~\ref{fig:tau-loss-sweep} separate two effects. Blue segments update only $W_1$ while keeping $W_2$ fixed, exposing the local $W_1$ optimization problem; red segments update only $W_2$ while keeping $W_1$ fixed, testing how easily the representation produced by $W_1$ can be used downstream.

This distinction supports the same mechanism as the two-phase experiment. Early blue segments often favor SGD, reflecting Muon's gradient-fidelity cost on the local subproblem. Muon's advantage appears more clearly once downstream adaptation has occurred: red segments can exploit the representation produced by the Muon trajectory, and the improved downstream layer can then make later $W_1$ subproblems easier. Detailed loss-gap and replay diagnostics are provided in Appendix~\ref{app:tau-diagnostics}. Overall, Muon may sacrifice immediate decrease on a layer-wise subproblem while producing representations that make downstream optimization easier.

\FloatBarrier
\section{Discussion}

\subsection{Implications for Optimizer Design}

The central design implication is that an optimizer need not be judged only by its immediate loss decrease on each layer-wise subproblem. An update can also shape the representation that downstream layers will later consume. From this perspective, Muon suggests a class of optimizers that explicitly balance \emph{gradient fidelity}, which favors local descent, and \emph{representation preservation}, which favors downstream usability. A natural direction is to make this balance adaptive, for example by controlling the strength of orthogonalization or spectral flattening according to layer spectra, conditioning, or downstream progress.

This view also suggests when Muon-like behavior should be most valuable. The payoff should be larger when downstream layers have enough capacity and training time to exploit a better-conditioned representation, and when optimization involves long chains of composed transformations. Conversely, when the main objective is nearly local, or when downstream adaptation is weak, the cost of reduced gradient fidelity may dominate.

\subsection{Limitations and Future Work}

Our experiments are deliberately controlled and linear. They demonstrate one possible mechanism by which Muon's slower local additive updates can be offset by faster downstream multiplicative adaptation, but they do not constitute a systematic study of the conditions under which this payoff appears in modern nonlinear networks. In particular, future work should test how the effect depends on depth, residual architecture, layer type, data imbalance, optimizer hyperparameters, and the amount of downstream adaptation available.

Another limitation is that our residual-connection analogy is mechanistic rather than architectural. Residual connections provide an option by reparameterizing the forward computation so that both $x_l$ and $\text{Block}(x_l)$ are explicitly available, whereas Muon provides a related option indirectly through the optimization trajectory. Muon does not insert an explicit skip path into the network, and real residual connections usually span an entire block rather than a single linear layer. Thus, the analogy is intended to highlight a shared option-like role for downstream computation, not to claim that the two mechanisms are interchangeable.

\section{Conclusion}

We presented a mechanistic interpretation of Muon as an implicit residual connection during optimization. In this view, orthogonalized updates trade some immediate gradient fidelity for representation preservation, making the resulting representations easier for downstream layers to use. Controlled two-phase and $\tau$-schedule linear experiments support this possibility: Muon can be slower on a local layer-wise target yet faster end-to-end because its learned representation improves downstream optimization. This perspective complements existing spectral, trust-region, curvature, and long-tail explanations, and suggests optimizer designs that reason explicitly about the payoff between local descent and downstream representation usability.

\FloatBarrier
\bibliography{iclr2026_conference}
\bibliographystyle{iclr2026_conference}

\appendix
\section{Detailed Existing Interpretations of Muon}
\label{app:existing-interpretations}

For completeness, we provide a more detailed summary of the existing interpretations of Muon discussed in the main text.

\subsection{Steepest Descent under the Spectral Norm}
\citet{bernstein2024old} interpret Muon as steepest descent under the matrix spectral norm. Let $G=\nabla f(W)$ and define \[ \operatorname{Orth}(G)=U V^\top, \qquad G=U\Sigma V^\top . \] The steepest descent direction under a norm $\|\cdot\|$ is obtained by solving \[ D \in \arg\min_{\|D\|\le 1} \langle G,D\rangle . \] When $\|\cdot\|$ is the matrix spectral norm $\|\cdot\|_2$, its dual norm is the nuclear norm $\|\cdot\|_*$. Therefore, \[ \min_{\|D\|_2\le 1}\langle G,D\rangle = -\max_{\|D\|_2\le 1}\langle G,D\rangle = -\|G\|_* . \] Using the singular value decomposition $G=U\Sigma V^\top$, the maximizer is $D=UV^\top=\operatorname{Orth}(G)$, and hence the steepest descent direction is \[ D^* = -\operatorname{Orth}(G). \] Equivalently, the associated non-Euclidean gradient direction, obtained from the duality mapping, is \[ G^\# = \|G\|_* \operatorname{Orth}(G), \] which leads to the update \[ W \leftarrow W - \eta \|G\|_* \operatorname{Orth}(G). \]

\subsection{Non-Euclidean Trust-Region Methods}

Rather than viewing Muon as steepest descent, \citet{kovalev2025trustregion} derive it exactly as the solution of a non-Euclidean trust-region subproblem,
\[
\min_{\Delta}
\;
\langle G,\Delta\rangle,
\qquad
\text{s.t.}\;
\|\Delta\|_2\le\eta,
\]
where $\|\cdot\|_2$ denotes the matrix spectral norm. Exploiting the duality between the spectral and nuclear norms, the optimal update admits the closed-form solution
\[
\Delta^*
=
-
\eta
\operatorname{Orth}(G),
\]
which coincides exactly with the Muon update
\[
W
\leftarrow
W
-
\eta
\operatorname{Orth}(G).
\]
Unlike the steepest-descent interpretation, this trust-region formulation recovers Muon without the additional nuclear-norm scaling factor. 

Although both interpretations identify the orthogonalized gradient as the update direction, they differ in the treatment of the step magnitude. The steepest-descent interpretation introduces an additional factor $\|G\|_*$ arising from the duality mapping of the spectral norm, whereas the trust-region formulation recovers the Muon update exactly with a fixed trust-region radius.

\subsection{Spectral-Norm-Constrained Updates with Weight Decay}

Recent work argues that Muon combined with decoupled weight decay should be viewed as optimizing under an implicit spectral-norm constraint~\citep{sfyraki2025lions}. Unlike the previous interpretations, which focus on the update direction of a single optimization step, this perspective emphasizes the constrained optimization trajectory rather than the update direction of a single step.

The central idea originates from the Frank--Wolfe algorithm, which maintains the iterates inside a prescribed convex set throughout optimization. At each iteration, Frank--Wolfe first solves the linear minimization oracle (LMO)
\[
S
=
\arg\min_{S\in\mathcal C}
\langle G,S\rangle,
\]
where $\mathcal C$ is the feasible set, and then moves the current iterate toward the selected extreme point,
\[
W
\leftarrow
(1-\gamma)W
+
\gamma S.
\]

When the feasible set is the spectral-norm ball
\[
\mathcal C
=
\{W:\|W\|_2\le r\},
\]
the solution of the LMO is
\[
S
=
-
r\,\operatorname{Orth}(G).
\]
Consequently, Frank--Wolfe naturally selects the orthogonalized gradient as the target point and updates the weight by interpolating between the current weight matrix and this extreme point.

Building on this observation, \citet{sfyraki2025lions} show that Muon with decoupled weight decay is exactly a stochastic Frank--Wolfe method on a spectral-norm ball, with weight decay keeping the trajectory inside the feasible region and the orthogonalized gradient selecting the extreme point to approach.
As a result, Muon is interpreted as optimizing under an implicit spectral-norm constraint rather than merely following an orthogonalized gradient direction at each step.

\subsection{Rare Directions and Long-Tail Learning}

Another line of work connects Muon to \emph{rare directions} in the gradient spectrum and to learning under heavy-tailed data.
By replacing the singular values of a matrix gradient with a common scale while preserving its singular vectors, orthogonalization prevents a few high-energy directions from dominating the update and gives relatively larger weight to weaker spectral components.
\citet{vasudeva2025spectral} study this mechanism systematically through spectral gradient descent (SpecGD), the canonical form of Muon in which each update is $UV^\top$ for a gradient $G=U\Sigma V^\top$.
They show that unlike Euclidean gradient descent, which learns dominant principal components of the data first, SpecGD learns all principal components at comparable rates.
On imbalanced classification problems, this spectral design yields better class-balanced loss and minority-group generalization early in training, and experiments with Muon and Shampoo confirm the same trend in practical settings.

This perspective has been extended to large language models through associative memory.
\citet{wang2025tailassociative} argue that Muon's advantage over Adam is concentrated in the associative-memory parameters of Transformers---notably the Value--Output attention weights and feed-forward networks---whose updates have an outer-product structure.
On heavy-tailed corpora, Muon produces a more isotropic singular spectrum than Adam and therefore learns tail classes more effectively, while matching Adam on frequent head classes; their one-layer associative-memory analysis further shows that Muon maintains balanced class-wise learning under imbalance, whereas Adam can exhibit large disparities depending on the representation embeddings.

\subsection{Curvature}

\citet{wang2026curvature} explain Muon's empirical advantage from the perspective of the local curvature of the loss landscape. Using a second-order Taylor expansion, they show that Muon and Adam achieve comparable first-order gains but differ substantially in their second-order curvature penalties. Specifically, the one-step loss decrease can be decomposed into a gradient-alignment term and a curvature term, and Muon consistently incurs a smaller curvature penalty. This advantage is further attributed to a lower \emph{Normalized Directional Sharpness} (NDS), rather than a smaller update norm. The paper further shows that data imbalance amplifies Muon's NDS advantage and proves, on stylized quadratic problems with heterogeneous curvature, that Muon's spectrally normalized updates distribute update energy more evenly across curvature modes, leading to lower average directional sharpness and faster optimization.

\subsection{Spectral Wasserstein Flow}

\citet{peyre2026spectralwasserstein} interpret Muon as the finite-particle limit of a continuous-time Spectral Wasserstein gradient flow. In the mean-field regime, optimization is formulated over probability measures equipped with a family of Spectral Wasserstein distances indexed by matrix norms. The trace norm recovers the classical quadratic Wasserstein geometry, the operator norm corresponds to Muon, and intermediate Schatten norms interpolate between the two. This interpretation establishes a unified optimal transport framework encompassing both Euclidean and spectrally normalized optimization.

\subsection{Alignment and Directional Descent Potential}

\citet{shumaylov2026muon} offer a complementary and partly skeptical view of explanations that attribute Muon's success to its exact orthogonalized spectrum. They argue that Muon's empirical advantage is better understood through two local quantities: batch gradient alignment, which measures how well the update direction agrees with individual example gradients, and directional descent potential, which measures how much descent can be obtained along the chosen direction. From this perspective, exact orthogonalization is not uniquely responsible for Muon's performance. Their proposed variants, including Freon and Kaon, modify the singular spectrum in random or inverted ways while retaining competitive behavior, suggesting that a broader class of spectrally shaped updates can preserve the relevant alignment--descent trade-off.

\section{Additional Experimental Details}

\subsection{Segment-wise Diagnostics of the $\tau$ Schedule}
\label{app:tau-diagnostics}

The colored segments in Figure~\ref{fig:tau-loss-sweep} separate two effects. Blue segments update only $W_1$ while keeping $W_2$ fixed, and therefore expose the local $W_1$ optimization problem. Red segments update only $W_2$ while keeping $W_1$ fixed, and therefore test how easily the representation produced by $W_1$ can be used by the downstream layer.

This distinction is important. In early blue segments, SGD often decreases the composition loss more rapidly than Muon, reflecting the gradient-fidelity cost discussed earlier. The advantage of Muon appears more clearly in the red segments: its $W_1$ trajectory can produce a representation that is easier for $W_2$ to exploit. After such red segments improve $W_2$, the resulting downstream layer also makes later blue-segment $W_1$ subproblems easier. This blue-slow, red-fast, easier-next-blue feedback loop can compensate for Muon's per-step cost on $W_1$ and still yield faster end-to-end convergence.

\begin{figure}[!htbp]
\centering
\includegraphics[width=\linewidth]{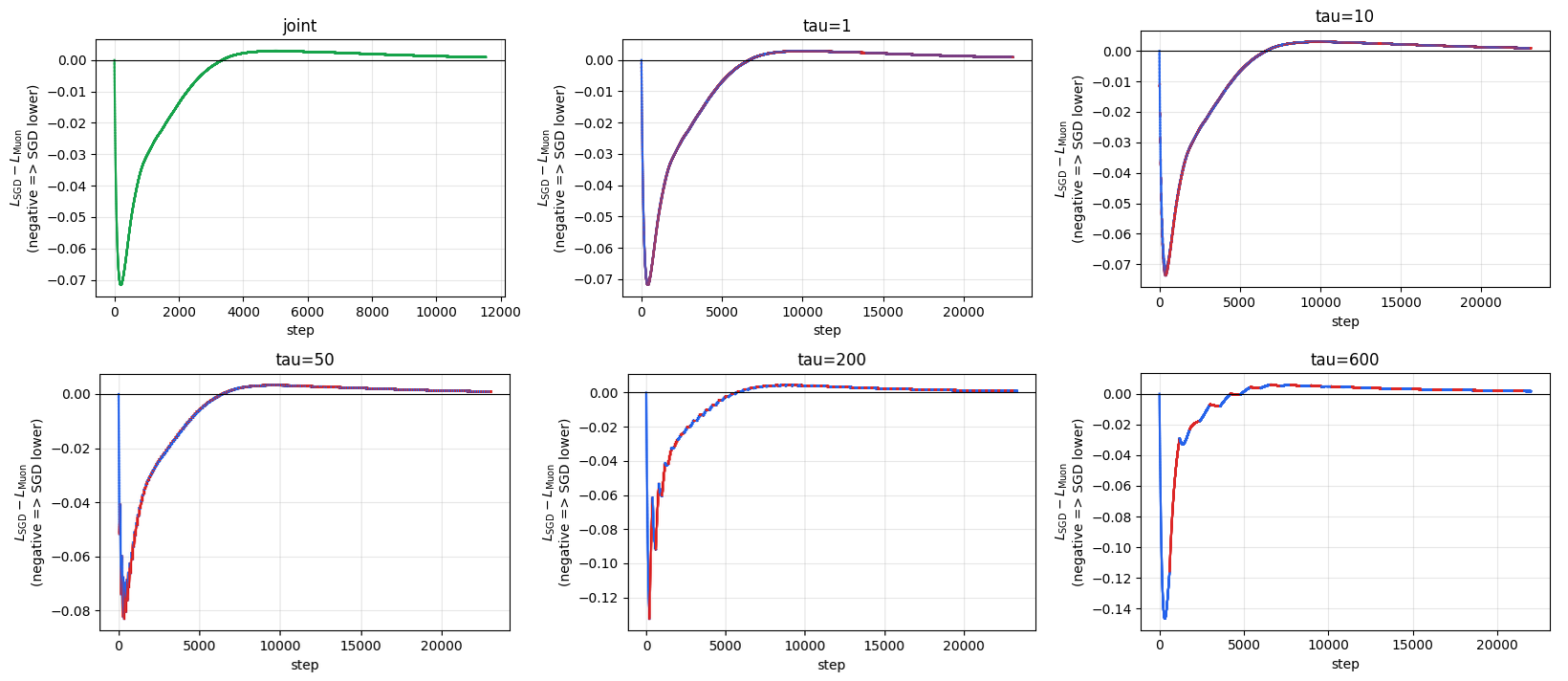}
\caption{Loss gap $L_{\rm SGD}-L_{\rm Muon}$ across the full $\tau$-schedule sweep. Negative values mean the SGD path has lower loss at the same step, while positive values mean the Muon path is ahead. The colored segments mark which layer is active.}
\label{fig:tau-gap-full}
\end{figure}

\begin{figure}[!htbp]
\centering
\includegraphics[width=\linewidth]{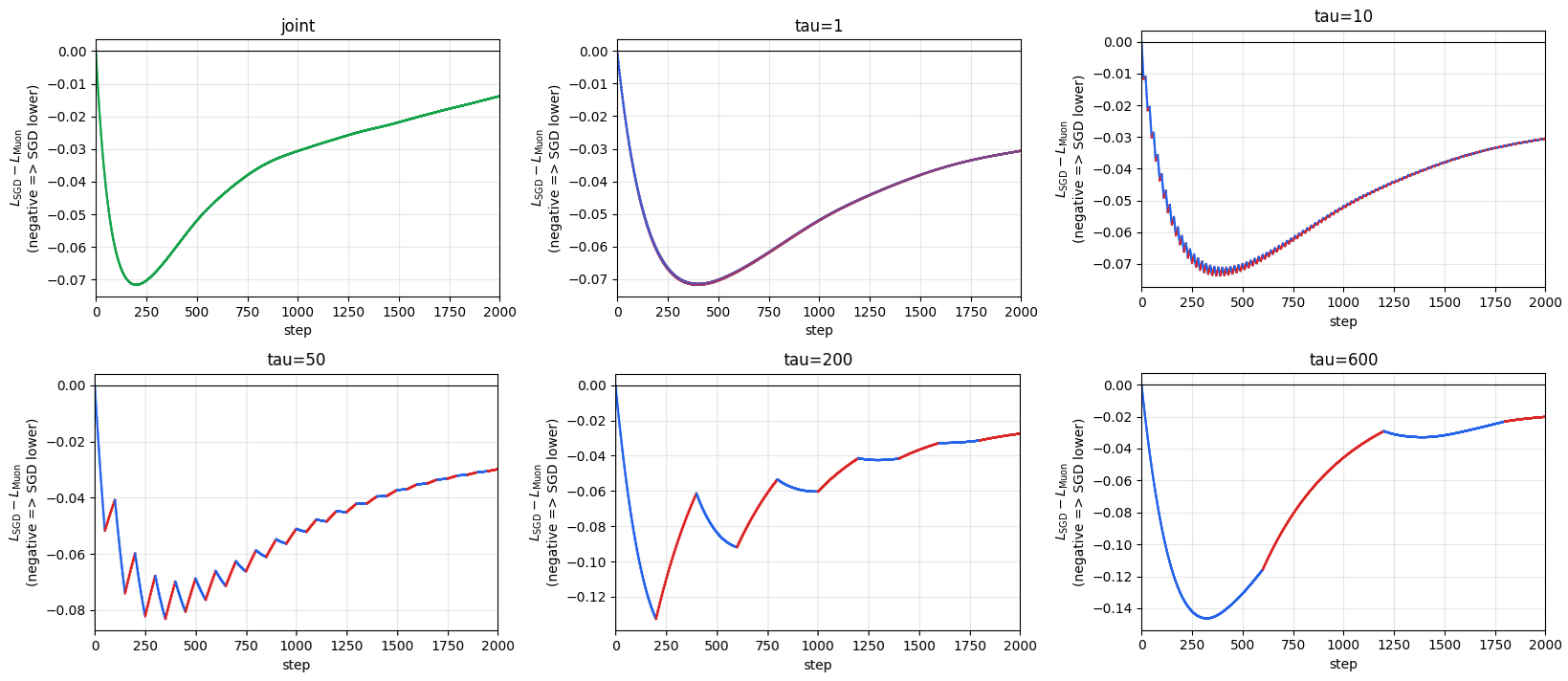}
\caption{Early-step zoom of the loss gap in Figure~\ref{fig:tau-gap-full}. The zoom highlights the initial local-optimization advantage of SGD on $W_1$ segments before Muon's representation advantage propagates through later $W_2$ updates.}
\label{fig:tau-gap-early}
\end{figure}

Figure~\ref{fig:tau-gap-full} plots this effect directly as the loss gap $L_{\rm SGD}-L_{\rm Muon}$, with Figure~\ref{fig:tau-gap-early} zooming in on the first 2000 steps. The early negative gaps show SGD's local advantage on fixed-$W_2$ blue segments, whereas later positive gaps show the accumulated benefit of the Muon trajectory once the downstream layer has adapted.

\begin{figure}[!htbp]
\centering
\includegraphics[width=0.92\linewidth]{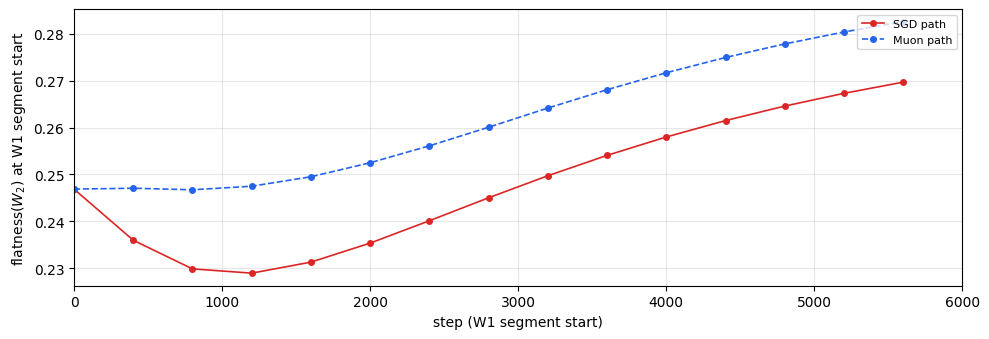}
\includegraphics[width=0.92\linewidth]{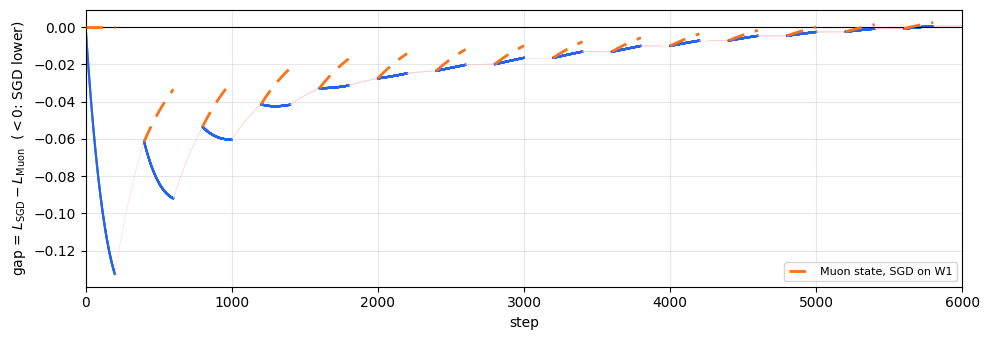}
\caption{Mechanism diagnostic for $\tau=200$. Top: spectral flatness of $W_2$ at the start of each $W_1$ segment (higher means a flatter singular spectrum). The Muon path tends to leave a flatter downstream layer before each blue segment. Bottom: loss gap $L_{\rm SGD}-L_{\rm Muon}$; blue portions are $W_1$-only segments and pale red portions are $W_2$-only segments. Negative gap means SGD has lower loss. The orange dashed replay, computed at each blue-segment start from the Muon-path $(W_1,W_2)$ with SGD used for $W_1$ in that segment only, tests whether Muon's mid-to-late catch-up is due to faster local optimization.}
\label{fig:tau200-mechanism}
\end{figure}

Why, then, can Muon appear to improve faster than SGD even inside mid-to-late blue segments, despite still orthogonalizing its updates? The answer is path-dependent rather than a claim that Muon has become the better local optimizer on an identical state. In the $\tau=200$ run, Figure~\ref{fig:tau200-mechanism} (top) records the spectral flatness of $W_2$ at the start of each blue segment, using the same effective-rank proxy as in Table~\ref{tab:two-phase-results}. The Muon path typically leaves a flatter $W_2$ spectrum before each blue segment; equivalently, the fixed-$W_2$ subproblem faced by $W_1$ is easier on the Muon trajectory. Muon can therefore gradually overtake SGD in later blue segments even while still sacrificing single-step speed to orthogonalization.

The bottom panel separates two sources of blue-segment behavior. When the gap decreases during a blue segment, SGD is reducing $L$ faster and widening its lead on the local $W_1$ subproblem, as expected from gradient fidelity. When the gap increases during a blue segment, Muon is catching up or pulling ahead even though only $W_1$ is being updated. The orange replay shows that this mid-to-late reversal is not because Muon has become locally superior on the same subproblem: if one fixes the Muon-path state at a blue-segment start and swaps Muon for SGD on $W_1$ in that segment alone, SGD can decrease the loss even faster on that \emph{same} subproblem. Muon's mid-to-late blue-segment gains therefore reflect an easier trajectory-induced subproblem created by earlier red and blue segments, not faster local optimization on an identical downstream state.

These observations extend the two-phase message to end-to-end training. Muon may sacrifice immediate decrease on a layer-wise subproblem, yet the representations produced along its trajectory can make downstream optimization easier, and that downstream improvement can feed back into the remaining upstream optimization.

\FloatBarrier

\subsection{Connection to Linear Probing}
\label{app:linear-probing}

The second phase of our experiment can be viewed as a simplified linear probing task. Instead of directly optimizing the matrix product $W_2W_1\rightarrow A$, consider a fixed encoder $W_1$ that produces the representation
\[
h=W_1x,
\]
while a linear probe $B$ is trained to recover a target transformation $M$. Depending on the choice of $M$, this includes both upstream reconstruction ($M=I$) and downstream prediction ($M=A$):
\[
L_M(B)
=
\frac12
\|BW_1-M\|_F^2.
\]

Let
\[
E_t
=
B_tW_1-M.
\]
A gradient descent step gives
\[
B_{t+1}
=
B_t-\eta(B_tW_1-M)W_1^\top,
\]
which implies
\[
E_{t+1}
=
E_t(I-\eta W_1^\top W_1).
\]

Therefore, the convergence rate is again completely determined by the spectrum of $W_1$. A flatter singular spectrum leads to faster convergence of the linear probe.

This establishes a direct connection between our two-phase linear model and the standard linear probing protocol widely used in representation learning. Although Muon may optimize the local objective more slowly, its tendency to produce better-conditioned representations can make them easier for downstream linear predictors to exploit.

\section{Proofs and Derivations}

\subsection{Phase~2 Convergence under a Fixed Representation}
\label{app:phase2-convergence}

Suppose
\[
W_1
=
U\Sigma V^\top.
\]
Then
\[
W_1^\top W_1
=
V\Sigma^2V^\top,
\]
and the error recursion
\[
E_{t+1}
=
E_t(I-\eta W_1^\top W_1)
\]
implies that each singular component of the error evolves independently as
\[
s_i(E_{t+1})
=
|1-\eta\sigma_i^2|
\,s_i(E_t),
\]
where $\sigma_i$ denotes the $i$-th singular value of $W_1$. Choosing
\[
\eta
=
\frac{\alpha}{\sigma_{\max}(W_1)^2},
\]
the slowest contraction factor becomes
\[
1-\frac{\alpha}{\kappa(W_1)^2},
\]
where $\kappa(W_1)$ is the condition number of $W_1$.

\end{document}